\documentclass{article}

\usepackage{arxiv}

\usepackage[utf8]{inputenc} 
\usepackage[T1]{fontenc}    
\usepackage{hyperref}       
\usepackage{url}            
\usepackage{booktabs}       
\usepackage{amsfonts}       
\usepackage{nicefrac}       
\usepackage{microtype}      
\usepackage{lipsum}
\usepackage{times}
\usepackage{epsfig}
\usepackage{graphicx}
\usepackage{amsmath}
\usepackage{amssymb}
\usepackage{rotating}
\usepackage{multirow}
\usepackage{subfig}
\usepackage{bm}
\usepackage{hhline}
\usepackage[norule,symbol,perpage]{footmisc}

\title{LivDet2023 - Fingerprint Liveness Detection Competition:\\Advancing Generalization}

\author{Marco Micheletto$^{1}$, Roberto Casula$^{1}$, Giulia Orrù$^{1}$, Simone Carta$^{1}$, Sara Concas$^{1}$,\\ \textbf{Simone Maurizio La Cava$^{1}$, Julian Fierrez$^{2}$, Gian Luca Marcialis$^{1}$}\\
$^{1}$  University of Cagliari, DIEE, Cagliari, Italy \\
$^{2}$ Universidad Autonoma de Madrid, BiDA Lab, Madrid, Spain\\
{\tt\small \{marco.micheletto, roberto.casula, giulia.orru, marcialis\}@unica.it, julian.fierrez@uam.es}
}
\usepackage{tikz}
\usepackage{textcomp}
\usepackage{hyperref}

\newcommand\copyrighttext{%
  \footnotesize \textcopyright 2023 IEEE. Personal use of this material is permitted.
  Permission from IEEE must be obtained for all other uses, in any current or future 
  media, including reprinting/republishing this material for advertising or promotional 
  purposes, creating new collective works, for resale or redistribution to servers or 
  lists, or reuse of any copyrighted component of this work in other works. 
  DOI: \href{}{}}
\newcommand\copyrightnotice{%
\begin{tikzpicture}[remember picture,overlay]
\node[anchor=south,yshift=10pt] at (current page.south) {\fbox{\parbox{\dimexpr\textwidth-\fboxsep-\fboxrule\relax}{\copyrighttext}}};
\end{tikzpicture}%
}
\begin{document}

\maketitle
\copyrightnotice
\begin{abstract}
The International Fingerprint Liveness Detection Competition (LivDet) is a biennial event that invites academic and industry participants to prove their advancements in Fingerprint Presentation Attack Detection (PAD). This edition, LivDet2023, proposed two challenges, ``Liveness Detection in Action'' and ``Fingerprint Representation'', to evaluate the efficacy of PAD embedded in verification systems and the effectiveness and compactness of feature sets. A third, ``hidden'' challenge is the inclusion of two subsets in the training set whose sensor information is unknown, testing participants' ability to generalize their models. Only \textit{bona fide} fingerprint samples were provided to participants, and the competition reports and assesses the performance of their algorithms suffering from this limitation in data availability.
\end{abstract}
\section{Introduction}

Due to their convenience and security, fingerprint-based authentication systems have garnered significant attention in numerous applications, ranging from financial transactions to healthcare management \cite{maltoni2022fingerprint}. However, these systems are vulnerable to various attacks, including spoofing or presentation attacks\cite{hadid15SPMspoofing}, where attackers use artificial replicas of live fingers to deceive the sensors \cite{Galbally_2009PRL}. Such attacks can lead to severe consequences, such as unauthorized access, identity theft, and financial fraud. To address this vulnerability, automated presentation attack detection (PAD) systems that utilize either hardware or software have been developed over the last few decades \cite{sousedik2014presentation, 2023_Book-PAD_Finger}.
Software-based methods, in particular, have seen significant advancements \cite{galbally14TIP} thanks to pattern recognition research innovations and larger datasets' availability \cite{2015_EncyBio_FpDB}. 
Among other initiatives, the International Fingerprint Liveness Detection Competition (LivDet)\footnote{https://sites.unica.it/livdet/} \cite{micheletto2023review}, now in its eighth edition, has significantly promoted research and development in this area since its inception in 2009 and has become a well-known benchmark for assessing the effectiveness of PAD techniques.
% ...

LivDet2023 presents two familiar challenges from the previous edition, ``Liveness Detection in Action'' and ``Fingerprint Representation'' \cite{micheletto2023review}, while introducing new evaluation criteria and datasets. Challenge 2 now evaluates system speed to encourage efficient and practical PAD systems that can operate in time-critical real-world scenarios. The retention of previous challenges ensures continuity and comparability with past results. 

Furthermore, LivDet2023 has set out to tackle a new challenge often overlooked in previous competition editions: the issue of generalization. Generalization \cite{chugh2019fingerprint} refers to the ability of systems to detect the authenticity of fingerprints in a wide variety of sensor technologies, Presentation Attack Instruments (PAIs) and attack scenarios rather than being limited to specific, predefined conditions.
Despite the improvements in the field, developing generalized PADs remains challenging for several reasons. Firstly, creating different PAIs for training is a non-trivial, expensive, and time-consuming task requiring skilled operators. Secondly, even with a comprehensive training dataset, existing methods often struggle to detect PAs captured using new acquisition methods or materials \cite{marasco2011robustness, micheletto2022mitigating}. Previous editions of LivDet have provided strong evidence of this phenomenon \cite{ micheletto2023review}.

Recent works have focused on addressing the lack of interoperability across fabrication materials by using one-class classification \cite{ ding2016ensemble, engelsma2019generalizing}. Unlike the traditional multi-class paradigm, one-class classification utilizes data from a single class, typically the bona fide class, for training the classifier. The ultimate objective of this approach is to establish a decision boundary around bona fide class samples that can accommodate as many samples as possible from that class while rejecting samples from other classes.
Driven by the potential of this approach, we have added a ``hidden'' challenge next to the previous ones. In particular, we included two subsets in the LivDet 2023 training set whose sensor information is unknown and contains only bona fide fingerprint samples. 
This challenge presents a valuable opportunity for participants to assess the effectiveness of their algorithms on unknown data, an essential aspect of real-world deployment. We hope this will increase LivDet2023's rigour and relevance while promoting the advancement of research and development in PADs. 

\section{LivDet2023}
As in the previous edition, two distinct challenges characterize the LivDet2023 competition:
\begin{itemize}
    \itemsep0em 
    \item Challenge 1, \textit{Liveness Detection in Action} \cite{6595860}: Competitors were asked to submit a complete algorithm capable of producing both the `` score'', that is, the probability of being a bona fide sample and the ``integrated score'', which combines the previous score with the probability of belonging to the claimed user. Participants in this challenge can choose whether to use the related ``user-specific'' information \cite{userspec}.
    \item Challenge 2, \textit{Fingerprint representation}: Compactness and discriminability of feature vectors are critical in modern authentication systems to ensure high performance in terms of accuracy and speed. With the aim of evaluating speed and compactness, we asked competitors to submit PADs that return the feature vector corresponding to the input image in addition to the score.
\end{itemize}
Furthermore, to evaluate the participating PADs' ability to generalize, an additional challenge, called \textit{Unknown sensors}, is introduced: in the training set, two sensors were unknown, the name and brand of the sensor were not declared, and only bona fide fingerprint samples were provided.
\subsection{Datasets and participants}
\label{sec:datasets}

Although the number of competitors was lower than in the previous edition, the competition showcased a diverse range of algorithms from each participant. It is worth noting that many competitors submitted multiple algorithms, which highlights their dedication to finding innovative solutions. Table \ref{table:Competitors} provides further details on each competitor, including the name of their presented algorithm, the type of solution adopted, and the challenge(s) in which they participated. 
In addition, we considered the quantity of data utilized by each participant in the training phase. In fact, some competitors have generated for each test set a model trained on data from the specific sensor (single, in Table \ref{table:Competitors}); others have generated a single model trained with data from multiple sensors suitable for multiple test sets (multiple, in Table \ref{table:Competitors}). Moreover, although we strongly advised utilizing only the LivDet 2023 dataset to maintain consistency in the results, some participants have employed additional data, which could have given them an advantage. Conversely, some competitors opted to use fewer data, typically omitting unknown sensors during the training phase. These instances will be designated by a plus (+) or minus (-) sign, respectively.

The LivDet 2023 training set and test set comprise four sub-sets containing fingerprint images from four different capture devices: GreenBit DactyScan 84C, Dermalog LF10, Jenetric LiveTouch Quattro and Integrated Biometrics Watson Mini. 

The sensors can be grouped into two categories: known and unknown sensors. GreenBit and Dermalog are known sensors; therefore, we provided competitors with these devices' names, brands, and technical details. The training set for these sensors included 25 users, for a total of 2750 images, subdivided into 1250 bona fide and 1500 PAs collected with the classic consensual method. On the other hand, Jenetric and Integrated Biometrics were unknown sensors: we did not declare any information about these devices to the competitors. The training set included only bona fide fingerprint samples, totalling 1250 samples.
To ensure the accuracy and dependability of the algorithms, our test sets were carefully designed to facilitate cross-material and cross-method experiments. In order to introduce more significant variability, we included synthetic fingerprints fabricated using materials different from those used in the training set. Furthermore, we incorporated presentation attacks generated with our semi-consensual ScreenSpoof technique \cite{casula2022towards}, which is known for its ability to produce highly realistic forgeries. The test set is four times larger than the training set, comprising 2500 bona fide and 6000 attack presentations (including both consensual and ScreenSpoof-generated PAIs) for known sensors. However, we deliberately included only 3000 ScreenSpoof-generated PAs for unknown sensors to create a more challenging scenario for the algorithms to detect and classify presentation attacks.

\begin{table*}[]
\centering
\caption{Device characteristics for LivDet 2023 datasets.}
\label{table:sensors}
\begin{tabular}{|c|c|c|c|c|c|} 
\hline
\textbf{Scanner} & \textbf{Model} & \textbf{Res. [dpi]}       & \textbf{Img Size}            & \textbf{Format}          & \textbf{Type}                 \\ 
\hline
Green Bit        & DactyScan84C   & 500                      & 500x500                      & JPEG                      & Optical                       \\ 
\hline
Dermalog         & LF10           & 500                      & 500x500                      & JPEG                     & Optical                       \\ 
\hline
Jenetric         &  Livetouch Quattro             & 500                      & 500x500                      & JPEG                      & Optical                       \\
\hline
Integrated Biometrics      &       Watson Mini         & 500                      & 500x500                      & JPEG                      & Hybrid                       \\\hline

\end{tabular}
\end{table*}

\begin{table*}[t]
\caption{Number of samples for each scanner and each part of the dataset.}
\label{table:datasetComposition}
\centering
\resizebox{\textwidth}{!}{%
\begin{tabular}{|c||c|c|c||c|c|c|c|c|c|}
\hline
& \multicolumn{3}{c||}{\textbf{Training}}      & \multicolumn{6}{c||}{\textbf{Test Consensual/ScreenSpoof}}  \\ \hline
\textbf{Dataset} & Bona fide & Latex\_v2 & RPro10 & Bona fide & BodyDouble\_new & ElmersGlue & R15 & GLS & Mix3  \\ \hline
Green Bit       &1250&750&750&2500& 1000/1000 & 1000/1000 & 1000/1000&-&-\\ \hline
Dermalog      &1250&750&750& 2500 & 1000/1000 &-& 1000/1000 & 1000/1000 & -\\ \hline \hline

Jenetric & 1250 & - & - & 2500 & 1500 & 1500 & - & - & \\ \hline
Int. Biometrics & 1250 & - & - & 2500 & - & 1500 & - & - & 1500 \\ \hline

\end{tabular}%

}
\end{table*}

\subsection{Algorithms submission}
\label{sec:algorit}
Algorithms for Challenge 1 must be submitted as console programs with the following parameters: \\\textit{[nameOfAlgorithm] [ndataset] [templateimagesfile] [probeimagesfile] [livenessoutputfile] [IMSoutputfile]}.\\ The parameter \textit{ndataset} is an identification number for the dataset used. The file \textit{templateimagesfile} contains a list of absolute paths to every template image stored in the system, while the file \textit{probeimagesfile} contains a list of absolute paths to each probe image that the algorithm will test.
The algorithm outputs are saved to the paths specified by the last two parameters. The file \textit{livenessoutputfile} contains the degree of ``liveness'' for each processed image, normalized between 0 and 100, where 100 indicates the highest degree of liveness, and 0 denotes a fake image. Fingerprint images with scores [0,50) are classified as ``presentation attack'', while those with scores [50,100] are classified as ``bona fide''.
The file \textit{IMSoutputfile} lists, for each probe image, the normalized probability of a fingerprint belonging to the declared identity and being authentic. Scores [0,50) classify the probe as ``presentation attack'' or the probe-template comparison as no-mated comparison, while scores [50,100] classify the comparison as bona-fide and mated.
The evaluation threshold is set to 50. If the algorithm is unable to process an image, the corresponding value in both outputs is set to -1000.

The submission process for Challenge 2 in LivDet 2023 is the same as in LivDet 2021. In addition to the parameters \textit{nameOfAlgorithm}, \textit{ndataset}, \textit{probeimagesfile}, and \textit{livenessoutputfile}, Challenge 2 applications require an additional parameter called \textit{embeddingsfile}, representing the file of feature vectors for each processed image.

\begin{table*}[]
\caption{Participants name and submitted algorithms, alongside information about their training approach. The terms 'Single' and 'Multiple' denote whether the model training used data from a specific or multiple sensors respectively. The symbols (+) and (-) indicate if participants used additional or fewer data for training their respective models.}
\label{table:Competitors}
\centering
\begin{tabular}{|c|c|c|c|c|c|} 
\hline
\textbf{Participant}                                                                                                      & \textbf{Algorithm name}                       & \textbf{Acronym} & \textbf{Challenge} & \textbf{Type}                   & \textbf{Training data}  \\ 
\hline
\multirow{2}{*}{\textbf{UNESP}}                                                                                           & Contreras\_1\_chl1/2\cite{9933421} & contr1           & 1,2                & \multirow{2}{*}{Hand-crafted}   & Single                  \\ 
\cline{2-4}\cline{6-6}         & Contreras\_2\_chl1/2\cite{9933421} & contr2           & 1,2                &                                 & Single                  \\ 
\hline
\multirow{6}{*}{\textbf{Peking University}}                                                                               & CIS\_PAD\_F                                   & CIS\_F           & 1                  & \multirow{15}{*}{Deep-learning} & Multiple(-)             \\ 
\cline{2-4}\cline{6-6}
                                                                                                                          & CIS\_PAD\_W                                   & CIS\_W           & 1                  &                                 & Multiple(-)             \\ 
\cline{2-4}\cline{6-6}
                                                                                                                          & CIS\_PAD\_W\_ensemble                         & CIS\_Wens        & 1                  &                                 & Multiple(-)             \\ 
\cline{2-4}\cline{6-6}
                                                                                                                          & CIS\_PAD\_F\_v2                               & CIS\_F2          & 1                  &                                 & Multiple(-)             \\ 
\cline{2-4}\cline{6-6}
                                                                                                                          & PAD\_Supcon\_cls                              & S\_cls           & 1                  &                                 & Multiple(-)             \\ 
\cline{2-4}\cline{6-6}
                                                                                                                          & PAD\_Supcon\_knn                              & S\_knn           & 1                  &                                 & Multiple                \\ 
\cline{1-4}\cline{6-6}
\textbf{Hanbat National University}                                                                                       & HNU\_AIM                                      & HNU              & 1                  &                                 & Multiple                \\ 
\cline{1-4}\cline{6-6}
\multirow{6}{*}{\begin{tabular}[c]{@{}c@{}}\textbf{Università degli Studi~}\\\textbf{di Napoli Federico II}\end{tabular}} & mod1                                          & unina1           & 1                  &                                 & Single/Multiple(-)      \\ 
\cline{2-4}\cline{6-6}
                                                                                                                          & mod2                                          & unina2           & 1,2                &                                 & Single/Multiple(-)      \\ 
\cline{2-4}\cline{6-6}
                                                                                                                          & mod3                                          & unina3           & 1,2                &                                 & Single/Multiple(-)      \\ 
\cline{2-4}\cline{6-6}
                                                                                                                          & unina\_grbt                                   & unina4           & 1                  &                                 & Multiple                \\ 
\cline{2-4}\cline{6-6}
                                                                                                                          & unina\_derm                                   & unina5           & 1                  &                                 & Multiple                \\ 
\cline{2-4}\cline{6-6}
                                                                                                                          & unina\_grbt\_derm                             & unina6           & 1                  &                                 & Multiple                \\ 
\cline{1-4}\cline{6-6}
\multirow{2}{*}{\textbf{JIIOV Technology}}                                                                                & run                                           & jiiov            & 1,2                &                                 & Multiple                \\ 
\cline{2-4}\cline{6-6}
                                                                                                                          & run\_all\_data                                & jiiov\_all       & 1,2                &                                 & Multiple(+)             \\
\hline
\end{tabular}
\end{table*}
\subsection{Performance Evaluation}
\label{sec:protocol}
In both challenges, the performance of the PADs will be evaluated using the standard PAD ISO metrics \cite{iso,iso1biomet}:
    \begin{itemize}
    \itemsep0em 
    \item PAD Accuracy: percentages of fingerprint images correctly classified by the PAD.
    \item BPCER (Bona fide Presentation Classification Error Rate): Rate of misclassified bona fide images.
    \item APCER (Attack Presentation Classification Error Rate): Rate of misclassified fake images.
    \end{itemize}
    
In Challenge 1, to evaluate the performance of the integrated system, we employed the following metrics:
\begin{itemize}
    \itemsep0em 
    \item FNMR (False Non-Match Rate): Rate of mated comparisons that result in rejection.
    \item FMR (False Match Rate): Rate of non-mated comparisons that result in acceptance.
    \item IAPAR (Impostor Attack Presentation Accept Rate): rate of presentation attacks that result in acceptance.
    \item Integrated Matching (IM) Accuracy: percentages of samples correctly classified by the integrated system.
\end{itemize}

To simulate real-world scenarios, we conducted comparisons using templates derived from bona fide fingerprints. The testing involved matching a fingerprint template to a fingerprint image from the same finger and user (mated), a fake fingerprint image from the same finger and user (presentation attack), or a bona fide fingerprint image from a different user (no-mated). Overall, we performed 5000 mated, 10000 no-mated and 10000 presentation attack comparisons.

Challenge 2 aimed to evaluate the compactness and the discriminability of feature vectors generated by various algorithms. We considered both the speed and size of the feature vectors to be essential parameters for this edition. To ensure fairness in the evaluation, we specified two machines where the algorithms were tested: a Desktop-PC Linux 18.04.1 Ubuntu or Windows 10 Pro system with an Intel® Core™ i9 9900K @ 3.60GHz processor, 64 GB DDR4 2.933 MHz RAM, and dual NVIDIA® GeForce® RTX 2080 Ti (11GB each) graphics cards. The final ranking was determined based on the speed of the algorithms in generating and comparing the feature vectors, their size, and the accuracy achieved on the specific dataset. The final score was obtained by combining the contributions of speed, compactness and PAD accuracy, normalized and averaged.

\section{Results}
\label{sec:Result}
This section examines the results of the algorithms submitted to LivDet2023. The global results of the two challenges are shown in  Tables \ref{tab:ch1overall} and \ref{tab:timech2}.

\begin{table}[tb]
\centering
\caption{Challenge 1: Integrated and PAD overall results. The jiiov\_all method is not considered in the final ranking as it uses additional data.}
\label{tab:ch1overall}
\begin{tabular}{|c|c|c|} 
\hline
\textbf{Algorithm} & \textbf{\begin{tabular}[c]{@{}c@{}} Overall PAD\\Accuracy [\%]\end{tabular}} & \textbf{\begin{tabular}[c]{@{}c@{}} Overall IM\\Accuracy [\%]\end{tabular}}    \\
\hline
Contr1             & 92,47                     & 53,69                      \\
Contr2             & 87,42                     & 42,09                      \\
CIS\_F             & 88,75                     & 91,55                      \\
CIS\_W             & 96,22                     & 95,99                      \\
\textbf{CIS\_Wens} & \textbf{97,54}            & \textbf{96,35}             \\
CIS\_F\_v2         & 89,22                     & 91,2                       \\
S\_cls             & 93,76                     & 94,11                      \\
S\_knn             & 94,05                     & 94,11                      \\
HNU\_AIM           & 75,8                      & 86,29                      \\
unina1             & 86,24                     & 93,14                      \\
unina2             & 87,42                     & 92,86                      \\
unina3             & 88,66                     & 93,47                      \\
unina4             & 87,23                     & 84,87                      \\
unina5             & 86,65                     & 93,07                      \\
unina6             & 88,3                      & 93,06                      \\
jiiov              & 90,12                     & 94,68                      \\ \hline\hline
jiiov\_all         & 87,11                     & 94,33                    \\ \hline
\end{tabular}
\end{table}

\begin{table*}[h]
\caption{Challenge 2: overall results.}
\label{tab:timech2}
\centering
\resizebox{0.7\textwidth}{!}{%
\begin{tabular}{|c||c|c|c|c|} 
\hline
\multicolumn{1}{|c||}{\textbf{Algorithm} }& \textbf{Overall Time/im [ms]}                   & \textbf{Feat. vect. size}                       & \textbf{Accuracy [\%]} & \textbf{Score}   \\ 

\hline
Contr1                                       & 1302.97 & 800 & \textbf{87.65}  &   0.57                  \\
Contr2                    & 4511.78    & 800 & 79.03      &         0.00     \\
unina2                    & 93.80      & \textbf{32} & 79.80 &  0.69   \\
unina3                    & 94.10 & \textbf{32} & 80.70      &      0.73            \\
\textbf{jiiov}                     & \textbf{46.89}  & 192 & 84.29     &       \textbf{0.80}         \\
jiiov\_all                & 47.42 & 192 & 80.55                &    0.66  \\ \hline

\end{tabular}
}
\end{table*}
Seventeen algorithms were submitted to Challenge 1, which evaluates the integration between PAD and comparator; the results are shown in Table \ref{tab:IMS_results_known} for the known, that is GreenBit and Dermalog, consensual test datasets, in Table \ref{tab:IMS_results_known_SS} for the known ScreenSpoof test datasets and in Table \ref{tab:IMS_results_unknown} for the two unknown sensors, Jenetric and Int. Biometrics.

Analyzing the results as a whole, it is evident that the test sets acquired with unknown sensors reported higher average errors. In particular, while on average, competitors achieved about 88\% IMS on data acquired on known sensors, this value dropped to 84\% for data from unknown sensors. Although, on average, this drop does not seem particularly significant, if we analyze the APCERs of the single methods, we can deduce that the rate of erroneously classified unknown presentation attacks is very high (14.91\% for known sensors vs. 39.58\% for unknown sensors).
This aspect shows that the interoperability problem in fingerprint presentation attack detection is still open.

In this challenge, the CIS\_W/Wens model emerges as the undisputed winner in terms of PAD/IM accuracy. This model, which is sensor-interoperable, has been designed to combine metric learning with the spoof detection task, aiming to encode more PAD-related information while minimizing sensor-related interference. Interestingly, despite the model's training being conducted with a smaller volume of data than what was fully available - specifically on the Dermalog and Greenbit sub-datasets - it does not appear that this necessarily led to superior performance. This suggests that the relationship between the volume of training data and the model's performance may not be directly proportional \cite{micheletto2022mitigating}.

Nevertheless, a notable observation from the data is the high FNMR, much more significant than typical verification systems without PAD \cite{maltoni2022fingerprint}. This distinctive characteristic applies across all participating algorithms and confirms what has been reported in \cite{micheletto2021fingerprint}, namely, the integration of a PAD algorithm has a substantial impact on the performance of the recognition system.

Compared to the previous year, there has been a shift in trend concerning the detection of consensual and ScreenSpoof attacks for the Greenbit dataset. In fact, the IAPAR is higher for this sensor in consensual scenarios. The reasons behind this phenomenon will be the subject of future research.

It is important to highlight that the handcrafted algorithms, Contr1 and Contr2, report a low IM accuracy due to the very high FMR. However, this behaviour is strictly linked to the choice of the comparator since PAD performances are in line with the other detectors. 

As we shift focus to deep learning methods, the underperformer is hnu\_aim. Despite its singular distinction as the quickest method in this edition (20 ms for probe/template comparison), it fails to demonstrate competitive potential in real-world applications due to its low accuracy.

The only algorithm that used additional data jiiov\_all has not demonstrated substantial effectiveness, particularly concerning PAD performance. It exhibited an unacceptably high error margin in the APCER metric, specifically when detecting ScreenSpoof-fabricated PAs.



For Challenge 2, six algorithms were submitted. The goal of this challenge is to encourage the development of algorithms that strike a balance between accuracy, speed, and compactness. These are crucial factors for ensuring high-performance fingerprint recognition.

Tables \ref{tab:ch2known} and \ref{tab:ch2unknown} present the results for known and unknown sensors. The overall evaluation considering processing time and feature vector size is shown in Table \ref{tab:timech2}. While the handcrafted methods exhibit the highest accuracy, they are also characterized by larger size and longer computational time. For example, the Contr2 method exceeds an average processing time of 4 seconds per image, which is impractical for real-world scenarios. 

Considering these aspects, the algorithm that offers the best compromise is jiiov. By leveraging the learning capabilities of a CNN, this algorithm effectively identifies patterns and distinctive characteristics in fingerprints, enabling fast image processing. In terms of compactness, the top-ranking algorithm is unina, which employs an autoencoder-based approach to achieve a condensed representation of relevant information.

However, it is important to note that the algorithms generally demonstrate a limited ability to handle the hidden challenge effectively. Despite reporting an acceptable BPCER, an average APCER close to 50\% is observed, implying that the PAs classification is akin to a coin toss. This result emphasizes the critical need to develop more sophisticated algorithms that can accurately identify and differentiate bona fide samples from presentation attacks, even without PA examples during training.

\begin{table*}[tb]\footnotesize
\caption{Challenge 1 Integrated and PAD Consensual results - Known scanners.}
\label{tab:IMS_results_known}
\centering
\resizebox{0.72\textwidth}{!}{%
\begin{tabular}{|c|c||c|c|c|c||c|c|c|}
\hline
 &  \textbf{Algorithms}           & \textbf{FNMR [\%]}  & \textbf{FMR [\%]}   & \textbf{IAPAR [\%]} & \textbf{\begin{tabular}[c]{@{}c@{}}IM\\Acc. [\%]\end{tabular}}& \textbf{BPCER [\%]}& \textbf{APCER [\%]}& \textbf{\begin{tabular}[c]{@{}c@{}}PAD\\Acc. [\%]\end{tabular}} \\ \hline
\multirow{17}{*}{\begin{sideways}\textbf{GreenBit  CC}\end{sideways}} & Contr1 & 1.42  & 98.5 & 22.82 & 51.19 & 1.36  & 22.9 & 90.03\\          
 & Contr2             & 0.94  & 97.6  & 43.54  & 43.36 & 0.41  & 43.76  & 82.25\\ 
& CIS\_F      & 15.32  & 0.1  & 8.98 & 93.30 & 0.16  & 11.45 & 95.33\\      
 & \textbf{CIS\_W}      & 15.24  & 0.12  & 5.53 & \textbf{94.69} & 0.05  & 7.22 & 97.08\\           
 & \textbf{CIS\_Wens}              & 15.24 & 0.12  & 5.53  & \textbf{94.69} & 0.05  & 7.22 & 97.08\\ 
 & CIS\_F\_v2 & 15.26 & 0.10 & 8.36 & 93.56 & 0.10 & 10.82 & 95.61 \\
 & S\_cls & 15.22 & 0.10 & 14.78 & 91.00 & 0.06 & 19.33 & 92.24 \\
 & S\_knn & 15.22 & 0.10 & 14.77 & 91.00 & 0.06 & 19.33 & 92.24 \\
 & HNU\_AIM                & 83.94  & 16.01  & 15.31  & 70.68 & 38.81  & 60.72 & 52.62\\          
 & unina1                 & 19.02  & 1.55  & 12.69  & 90.50 & 0.00  & 32.86 & 86.86\\          
 & unina2                   & 23.32 & 1.51 & 10.38 & 90.58 & 5.35 & 26.51 & 86.19\\
 & unina3 & 21.46 & 1.54 & 11.37 & 90.54 & 2.61 & 30.67 & 86.16\\
 & unina4 & 20.92 & 3.54 & 13.23 & 89.11 & 2.01 & 32.75 & 85.70 \\
 & unina5 & 19.02 & 1.55 & 12.90 & 90.42 & 0.00 & 33.07 & 86.77 \\
 & unina6 & 19.02 & 1.55 & 12.90 & 90.42 & 0.00 & 33.07 & 86.77 \\ 
 & jiiov             & 18.06  & 0.09  & 8.75  & 92.85 & 2.75  & 28.56  & 86.93\\           
 & jiiov\_all                 & 16.38 & 0.08 & 5.55 & 94.47 & 1.08  & 16.83 & 92.62\\\hline
\multirow{17}{*}{\begin{sideways}\textbf{Dermalog  CC}\end{sideways}} & Contr1 & 3.00  & 97.44 & 3.57 & 59.00 & 1.58  & 3.57 & 97.62\\          
 & Contr2             & 1.86  & 97.38  & 6.04  & 58.26 & 1.83  & 6.04 & 96.48\\   
& CIS\_F      & 14.86  & 0.01  & 2.52  & 96.02 & 0.14  & 3.65  & 98.46\\        
 & \textbf{CIS\_W}     & 14.90  & 0.03  & 0.02  &   \textbf{97.00} & 0.20  & 0.06  & 99.86\\   
 & \textbf{CIS\_Wens }             & 14.90  & 0.03  & 0.02  & \textbf{97.00} & 0.20  & 0.06  & 99.86\\   
 & CIS\_F\_v2 & 14.82 & 0.01 & 1.56 & 96.41 & 0.06 & 0.29 & 99.08 \\
 & S\_cls & 14.80 & 0.00 & 0.18 & 96.97 & 0.06 & 0.29 & 99.85 \\
 & S\_knn & 14.80 & 0.00 & 0.18 & 96.97 & 0.06 & 0.29 & 99.85 \\
& HNU\_AIM                 & 37.58 & 7.87  & 0.27  & 89.23 & 0.39  & 3.09  & 98.53\\          
 & unina1                 & 18.78 & 2.71  & 2.39  & 94.20 & 0.03  & 5.06  & 97.96\\          
 & unina2                   & 21.08 & 2.59 & 5.49 & 92.55 & 3.34 & 9.99 & 94.00\\
 & unina3 & 19.34 & 2.69 & 1.08 & 94.62 & 0.70 & 2.10 & 98.74\\
 & unina4 & 20.06 & 4.74 & 3.52 & 92.68 & 1.92 & 5.20 & 96.77 \\
 & unina5 & 18.78 & 2.71 & 2.53 & 94.15 & 0.03 & 5.20 & 97.90 \\
 & unina6 & 18.78 & 2.71 & 2.53 & 94.15 & 0.03 & 5.20 & 97.90 \\
  & jiiov             & 18.32  & 0.12  & 0.34  & 96.15 & 3.19  & 1.59  & 97.45\\          
 & jiiov\_all                 & 16.10 & 0.17  & 0.18  & 96.64 & 0.95  & 1.31  & 98.90\\    \hline

 \end{tabular}}
\end{table*}
\begin{table*}[tb]\footnotesize
\caption{Challenge 1 Integrated and PAD ScreenSpoof results - Known scanners.}
\label{tab:IMS_results_known_SS}
\centering
\resizebox{0.72\textwidth}{!}{%
\begin{tabular}{|c|c||c|c|c|c||c|c|c|}
\hline
 &   \textbf{Algorithms}           & \textbf{FNMR [\%]}  & \textbf{FMR [\%]}   & \textbf{IAPAR [\%]} & \textbf{\begin{tabular}[c]{@{}c@{}}IM\\Acc. [\%]\end{tabular}}& \textbf{BPCER [\%]}& \textbf{APCER [\%]}& \textbf{\begin{tabular}[c]{@{}c@{}}PAD\\Acc. [\%]\end{tabular}} \\ \hline
\multirow{17}{*}{\begin{sideways}\textbf{GreenBit  SS}\end{sideways}} & Contr1 & 1.42  & 98.79 & 1.52 & 59.59 & 1.17  & 1.52 & 98.69\\          
 & Contr2             & 0.94  & 97.45  & 3.55 & 59.41 & 0.45 & 6.39 & 97,18\\ 
& CIS\_F      & 15.32  & 0.03  & 6.91 & 94.16 & 0.12  & 14.45 & 94.15\\          
 & \textbf{CIS\_W}      & 15.24  & 0.07 & 0.18  & \textbf{96.85} & 0.03  & 0.48  & 99.79\\          
 & \textbf{CIS\_Wens}               & 15.24  & 0.07 & 0.18  & \textbf{96.85} & 0.03  & 0.48  & 99.79\\   
 & CIS\_F\_v2 & 15.26 &0.03 & 9.10 & 94.90 & 0.09 & 15.65 & 93.68 \\
 & S\_cls & 15.22 & 0.03 & 13.59 & 91.51 & 0.03 & 28.69 & 88.50 \\
 & S\_knn & 15.22 & 0.03 & 13.59 & 91.51 & 0.03 & 28.69 & 88.50 \\
& HNU\_AIM                 & 28.40  & 14.20  & 6.07 & 82.21 & 0.23  & 65.85 & 73.52\\          
 & unina1                 & 19.02  & 1.73  & 4.62 & 93.66 & 0.00  & 40.40 & 83.84\\          
 & unina2                   & 23.32 & 1.69 & 0.04 & 94.64 & 5.23 & 0.03 & 96.85\\
 & unina3 & 21.46 & 1.71 & 0.32 & 94.90 & 2.59 & 7.39 & 95.49\\
 & unina4 & 26.04 & 9.33 & 39.11 & 75.42 & 7.69 & 28.17 & 84.12 \\
 & unina5 & 19.02 & 1.73 & 4.93 & 93.53 & 0.00 & 40.71 & 83.72 \\
 & unina6 & 19.02 & 1.73 & 4.93 & 93.53 & 0.00 & 40.71 & 83.72 \\
 & jiiov             & 18.06  & 0.11  & 1.02 & 95.94 & 2.68  & 4.81 & 96.47\\          
 & jioov\_all                 & 16.38 & 0.11 & 7.42 & 96.24 & 0.99  & 38.39 & 84.05\\\hline
\multirow{17}{*}{\begin{sideways}\textbf{Dermalog SS}\end{sideways}}  & Contr1 & 3.00  & 97.32 & 1.66 & 59.81 & 1.61  & 1.66 & 98.37\\          
 & Contr2             & 1.86  & 97.11  & 3.12 & 39.91 & 2.09  & 3.40 & 97.39\\ 
& CIS\_F     & 14.86  & 0.07  & 3.03 & 95.79 & 0.18  & 6.39 & 97.34\\          
 & \textbf{CIS\_W}     & 14.9  & 0.05  & 0.42 & \textbf{96.83} & 0.23  & 0.64 & 99.61\\         
 & \textbf{CIS\_Wens }             & 14.9  & 0.05  & 0.42 & \textbf{96.83} & 0.23  & 0.64 & 99.61\\ 
 & CIS\_F\_v2 & 14.82 & 0.07 & 16.49 & 90.41 & 0.16 & 25.63 & 89.65 \\
 & S\_cls & 14.80 & 0.07 & 6.35 & 94.47 & 0.08 & 12.19 & 95.08 \\
 & S\_knn & 14.80 & 0.07 & 6.36 & 94.47 & 0.08 & 12.19 & 95.08 \\
& HNU\_AIM                 & 37.58 & 7.68  & 2.03 & 88.60 & 0.46  & 27.30 & 88.84\\          
 & unina1                 & 18.78 & 2.91  & 1.38 & 94.53 & 0.06  & 8.09 & 96.73\\         
 & unina2                   & 21.08 & 2.85 & 0.05 & 94.62 & 3.35 & 0.62 & 97.74\\
 & unina3 & 19.34 & 2.90 & 0.18 & 94.90 & 0.69 & 1.01 & 99.18\\
 & unina4 & 18.98 & 3.16 & 4.83 & 93.01 & 0.31 & 8.18 & 96.54 \\
 & unina5 & 18.78 & 2.94 & 1.64 & 94.41 & 0.06 & 8.35 & 96.62 \\
 & unina6 & 18.78 & 2.94 & 1.64 & 94.41 & 0.06 & 8.35 & 96.62 \\ 
 & jioov             & 18.32  & 0.06  & 1.18 & 95.84 & 3.38  & 3.48 & 96.58\\          
 & jiiov\_all                 & 16.10 & 0.13  & 8.02 & 93.52 & 1.04  & 31.34 & 86.84\\\hline
 \end{tabular}}
\end{table*}

\begin{table*}[tb]\footnotesize
\caption{Challenge 1 Integrated and PAD results - Unknown scanners.}
\label{tab:IMS_results_unknown}
\centering
\resizebox{0.72\textwidth}{!}{%
\begin{tabular}{|c|c||c|c|c|c||c|c|c|}
\hline
 &  \textbf{Algorithms}           & \textbf{FNMR [\%]}  & \textbf{FMR [\%]}   & \textbf{IAPAR [\%]} & \textbf{\begin{tabular}[c]{@{}c@{}}IM\\Acc. [\%]\end{tabular}}& \textbf{BPCER [\%]}& \textbf{APCER [\%]}& \textbf{\begin{tabular}[c]{@{}c@{}}PAD\\Acc. [\%]\end{tabular}} \\  \hline
 \multirow{17}{*}{\begin{sideways}\textbf{Jenetric SS}\end{sideways}} & Contr1 & 1,38  & 98.45 & 21.99 & 51.55 & 0.79 & 22.17  & 90.66\\          
 & Contr2             & 0.00  & 99.99  & 50.55 & 39.78 & 0.00  & 51.21 & 79.52\\  
 & CIS\_F      & 21.76  & 0.04  & 7.06 & 92.81 & 4.75  & 16.71 & 90.47\\          
 & CIS\_W      & 26.48 & 0.05  & 0.00 & 94.68 & 10.41  & 0.00 & 93.75\\         
 & \textbf{CIS\_Wens}              & 20.76  & 0.05  & 0.75 & \textbf{95.53} & 3.39  & 1.56 & 97.34\\ 
 & CIS\_F\_v2 & 19.92 & 0.04 & 18.34 & 88.66 & 2.37 & 32.46 & 85.60 \\
 & S\_cls & 20.46 & 0.04 & 6.38 & 93.34 & 3.28 & 13.80 & 92.51 \\
 & S\_knn & 18.86 & 0.04 & 10.44 & 92.04 & 1.55 & 16.71 & 92.38 \\
 & HNU\_AIM                 & 18.66 & 9.54  & 1.60 & 91.81 & 0.11  & 59.89 & 75.98\\          
 & unina1                 & 24.24 & 2.14  & 1.84 & 93.56 & 5.71  & 11.37 & 92.03\\         
 & unina2               & 23.58 & 2.03 & 5.31  & 92.35 & 4.53 & 18.82 & 89.75\\
 & unina3          & 21.36 & 2.18 & 3.47 & 93.47 & 2.00 & 16.11 & 92.36\\
 & unina4 & 30.50 & 12.66 & 40.40 & 72.68 & 11.25 & 34.39 & 79.49 \\
 & unina5 & 20.40 & 2.33 & 3.56 & 93.56 & 1.01 & 19.18 & 91.72 \\
 & unina6 & 20.04 & 2.24 & 6.40 & 92.54 & 0.04 & 35.59 & 85.74 \\
  & jiiov             & 19.92  & 0.09  & 12.06 & 91.16 & 0.32  & 52.09 & 78.97\\          
 & jiiov\_all                 & 20.92 & 0.11  & 6.90 & 93.01 & 2.17  & 30.47 & 86.51\\  \hline

 \multirow{17}{*}{\begin{sideways}\textbf{Int. Biometrics SS}\end{sideways}}  & Contr1 & 3.04 & 97.09 & 48.87 & 41.01 & 1.45 & 49.16  & 79.46\\          
 & Contr2             & 0.00  & 99.75 & 70.71 & 11.82 & 0.00  & 70.77 & 71.69\\  
 & CIS\_F      & 16.72  & 0.01  & 48.62 & 77.20 & 17.03 & 82.64 & 56.73\\          
 & CIS\_W      & 19.38 & 0.03  & 0.52 & 95.90 & 20.31 & 1.54 & 87.20\\         
 & CIS\_Wens              & 10.24  & 0.03  & 1.87 & 97.19 & 11.13 & 4.38 & 91.57\\          
 & CIS\_F\_v2 & 5.26 & 0.02 & 35.17 & 84.87 & 5.00 & 63.27 & 71.69 \\ 
 & S\_cls & 3.84 & 0.02 & 4.68 & 97.35 & 3.70 & 8.50 & 94.38 \\
 & \textbf{S\_knn} & 5.70 & 0.02 & 0.45 & \textbf{98.67} & 5.68 & 0.85 & 96.25 \\
 & HNU\_AIM                 & 3.98 & 16.13  & 3.86 & 91.21 & 0.00  & 86.23 & 65.51\\          
 & unina1                 & 3.90 & 3.32  & 13.68 & 92.42 & 0.00  & 100.00 & 60.00\\         
 & unina2               & 3.94 & 3.32 & 13.68 & 92.41 & 0.05 & 99.96 & 59.99\\
 & unina3         & 3.94 & 3.32 & 13.68 & 92.41 & 0.00 & 99.96 & 60.02\\
 & unina4 & 11.04 & 8.29 & 20.42 & 86.31 & 8.70 & 35.13 & 80.73 \\
 & unina5 & 4.34 & 3.25 & 13.72 & 92.34 & 0.72 & 90.95 & 63.19 \\
 & unina6 & 6.92  & 3.41 & 9.84 & 93.32 & 4.21 & 46.09 & 79.04 \\
 & jiiov             & 1.94  & 0.15  & 8.46 & 96.17 & 1.08  & 37.55 & 84.33\\          
 & jiiov\_all                 & 1.30 & 0.15  & 12.59 & 94.64 & 1.11  & 36.06 & 73.76\\ \hline
\end{tabular}}
\end{table*}

\begin{table*}[h]
\caption{Challenge 2 PAD accuracy of the algorithms on the test sets. For each known dataset  the rate of misclassified bona fide  and fake fingerprints are reported. The last column is relative to the average of the total accuracy on the four known datasets.}
\label{tab:ch2known}
\centering
\resizebox{0.95\textwidth}{!}{%
\begin{tabular}{|c||c|c|c|c|c||c|c|c|c|c||c|} 
\hline
\multicolumn{1}{|c||}{\multirow{3}{*}{\textbf{Algorithm} }} & \multicolumn{5}{c||}{\textbf{Green Bit~} }                  & \multicolumn{5}{c||}{\textbf{Dermalog} }                      & \multirow{3}{*}{\begin{tabular}[c]{@{}c@{}}\textbf{Overall}\\\textbf{PAD}\\\textbf{Acc. [\%]} \end{tabular}}  \\ 
\cline{2-11}
\multicolumn{1}{|c||}{}       & \multicolumn{1}{c|}{\multirow{2}{*}{\begin{tabular}[c]{@{}c@{}} \textit{BPCER}\\\textit{ [\%]} \end{tabular}}} & \multicolumn{2}{c|}{CC}          & \multicolumn{2}{c||}{SS}            & \multicolumn{1}{c|}{\multirow{2}{*}{\begin{tabular}[c]{@{}c@{}} \textit{BPCER}\\\textit{ [\%]} \end{tabular}}} & \multicolumn{2}{c|}{CC}                     & \multicolumn{2}{c||}{SS}   &    \\ 
\cline{3-6}\cline{8-11}
\multicolumn{1}{|c||}{}       & \multicolumn{1}{c|}{}                              & \begin{tabular}[c]{@{}c@{}}\textit{APCER}\\\textit{ [\%]} \end{tabular} & \begin{tabular}[c]{@{}c@{}}\textit{PAD }\\\textit{ Acc. [\%]} \end{tabular} & \begin{tabular}[c]{@{}c@{}}\textit{APCER}\\\textit{ [\%]}\end{tabular} & \begin{tabular}[c]{@{}c@{}}\textit{PAD}\\\textit{ Acc. [\%]}\end{tabular} & \multicolumn{1}{c|}{}& \begin{tabular}[c]{@{}c@{}}\textit{APCER}\\\textit{ [\%]} \end{tabular} & \begin{tabular}[c]{@{}c@{}}\textit{PAD}\\\textit{ Acc. [\%]} \end{tabular} & \begin{tabular}[c]{@{}c@{}}\textit{APCER}\\\textit{ [\%]} \end{tabular} & \begin{tabular}[c]{@{}c@{}}\textit{PAD }\\\textit{ Acc. [\%]} \end{tabular} &                           \\ 
\hline
Contr1                        & 1.20               & 23.13       & 86.84               & 1.57      & \textbf{98.60}             & 1.64                & 3.43       & 97.38              & 1.70       & 98.33                & \textbf{95.29}                   \\
Contr2                   & 0.44                 & 43.83       & 75.89               & 6.13      & 96.45             & 2.12                 & 5.87        & 95.84              & 3.53       & 97.11               & 91.32                   \\
unina2                  & 4.96                      & 26.57       & 83.25               & 0.03      & 97.73             & 3.44                 & 9.97        & 93.00              & 3.44       & 98.07               & 93.01                 \\
unina3                     & 2.32                      & 31.23       & 81.91               & 7.60      & 94.80             & 0.68                 & 2.17        & 98.51              & 1.07       & \textbf{99.11}               & 93.58                   \\
jiiov                 & 2.68                & 28.50       & 83.24               & 4.63      & 96.25             & 3.60                 & 1.63       & 97.47              & 3.47       & 96.47               & 93.36                   \\
jiiov\_all                    & 0.96                 & 17.03        & \textbf{90.27}               & 38.47      & 78.58             & 1.12                 & 1.30        & \textbf{98.78}              & 31.00       & 82.58               & 87.55                   \\ \hline

\end{tabular}
}
\end{table*}

\begin{table*}[h]
\caption{Challenge 2 PAD accuracy of the algorithms on the test sets. For each unknown dataset  the rate of misclassified bona fide  and fake fingerprints are reported. The last column is relative to the average of the total accuracy on the four unknown datasets.}
\label{tab:ch2unknown}
\centering
\resizebox{0.95\textwidth}{!}{%
\begin{tabular}{|c||c|c|c||c|c|c||c|} 
\hline
\multicolumn{1}{|c||}{\multirow{3}{*}{\textbf{Algorithm} }} & \multicolumn{3}{c||}{\textbf{Jenetric} }                  & \multicolumn{3}{c||}{\textbf{Integrated Biometrics} }                      & \multirow{2}{*}{\begin{tabular}[c]{@{}c@{}}\textbf{Overall PAD}\\\textbf{Acc. [\%]} \end{tabular}}  \\ 
\cline{2-7}
\multicolumn{1}{|c||}{}       &  \textit{BPCER [\%]}    & \textit{APCER [\%]}   & \textit{PAD Acc. [\%]}    & \textit{BPCER [\%]}    & \textit{APCER [\%]}   & \textit{ Acc. [\%]}  &  \\ 
  
\hline
Contr1                        & 0.84               & 22.33       & 87.44               & 1.48      & 49.07             & 72.56                & \textbf{80.00}                       \\
Contr2                   & 0.00                 & 51.23       & 72.05               & 0.00      & 70.77             & 61.40                 & 66.73                        \\
unina2                  & 4.36                      & 18.90       & 87.71               & 0.04      & 99.97             & 45.45                 & 66.58          \\
unina3                     & 1.88                      & 16.50       & \textbf{90.15}               & 0.00      & 99.97             & 45.47                 & 67.81                        \\
jiiov                 & 0.36                & 52.13       & 71.40               & 1.08      & 37.53             & \textbf{79.03}                 & 75.22                        \\
jiiov\_all                    & 2.40                 & 30.13        & 82.47               & 1.08      & 63.83             & 64.60                & 73.54                      \\ \hline

\end{tabular}
}
\end{table*}
\section{Discussion and conclusions}

The eighth edition of the Fingerprint Liveness Detection Competition allowed for the evaluation of the degree of interoperability of current PADs, in addition to the impact of integrating a PAD system with an AFIS and the level of compactness, speed, and representativeness.
To simulate a worst-case scenario for an AFIS designer, we only provided competitors with bona-fide samples for two of the sensors used. The competitors have faced the challenge in the most different ways: someone has trained with only the data of the other sensors to carry out a sort of transfer domain; others have used data from multiple sensors to increase the PAD's ability to generalize; others have used information from the training data to generate unknown PAs synthetically. No one reported using a one-class classifier.
Although the proposed solutions were very different, the APCER on the unknown sensors is still high, especially on one of the two sensors. This shows that the interoperability problem is still open but that solutions to solve it are under development and have potential.
A comparison with past LivDet editions reveals a pause in the rise of accuracy typical of earlier editions, with some fluctuations due to the diverse challenges and materials involved.

\section*{Acknowledgements}
We would like to thank Dirk Morgeneier and his company Jenetric for sponsoring the LivDet 2023 competition.\\
J.F. is supported by project BBforTAI (PID2021-127641OB-I00MICINN/FEDER).

\newpage
{\small
\bibliographystyle{abbrv}
\bibliography{egbib}
}

\end{document}